\documentclass[10pt,twocolumn,letterpaper]{article}

\usepackage[pagenumbers]{iccv} 

\usepackage{cuted}
\usepackage{xcolor}
\definecolor{darkgreen}{RGB}{0,128,0}

\definecolor{iccvblue}{rgb}{0.21,0.49,0.74}
\usepackage[pagebackref,breaklinks,colorlinks,allcolors=iccvblue]{hyperref}

\usepackage{pifont}

\def\paperID{12820} 
\def\confName{ICCV}
\def\confYear{2025}

\title{\vspace{-12pt}MoCha: Towards Movie-Grade Talking Character Synthesis}

\author{
Cong Wei\textsuperscript{1,2},
Bo Sun\textsuperscript{2},
Haoyu Ma\textsuperscript{2},  
Ji Hou\textsuperscript{2},  
Felix Juefei-Xu\textsuperscript{2},  
Zecheng He\textsuperscript{2}, 
Xiaoliang Dai\textsuperscript{2}, \\
Luxin Zhang\textsuperscript{2},
Kunpeng Li\textsuperscript{2},
Tingbo Hou\textsuperscript{2},  
Animesh Sinha\textsuperscript{2},  
Peter Vajda\textsuperscript{2},  
Wenhu Chen\textsuperscript{1} \\
\textsuperscript{1}University of Waterloo,  
\textsuperscript{2}GenAI, Meta\\
\url{https://congwei1230.github.io/MoCha}
}

\newcommand{\MoCha}{MoCha\xspace}
\newcommand{\MoChaBench}{MoCha-Bench\xspace}

\begin{document}

\twocolumn[{
\maketitle
\renewcommand\twocolumn[1][]{#1}
\begin{center}
    \captionsetup{type=figure}
    \vspace{-5pt}
    \includegraphics[width=0.85\linewidth]{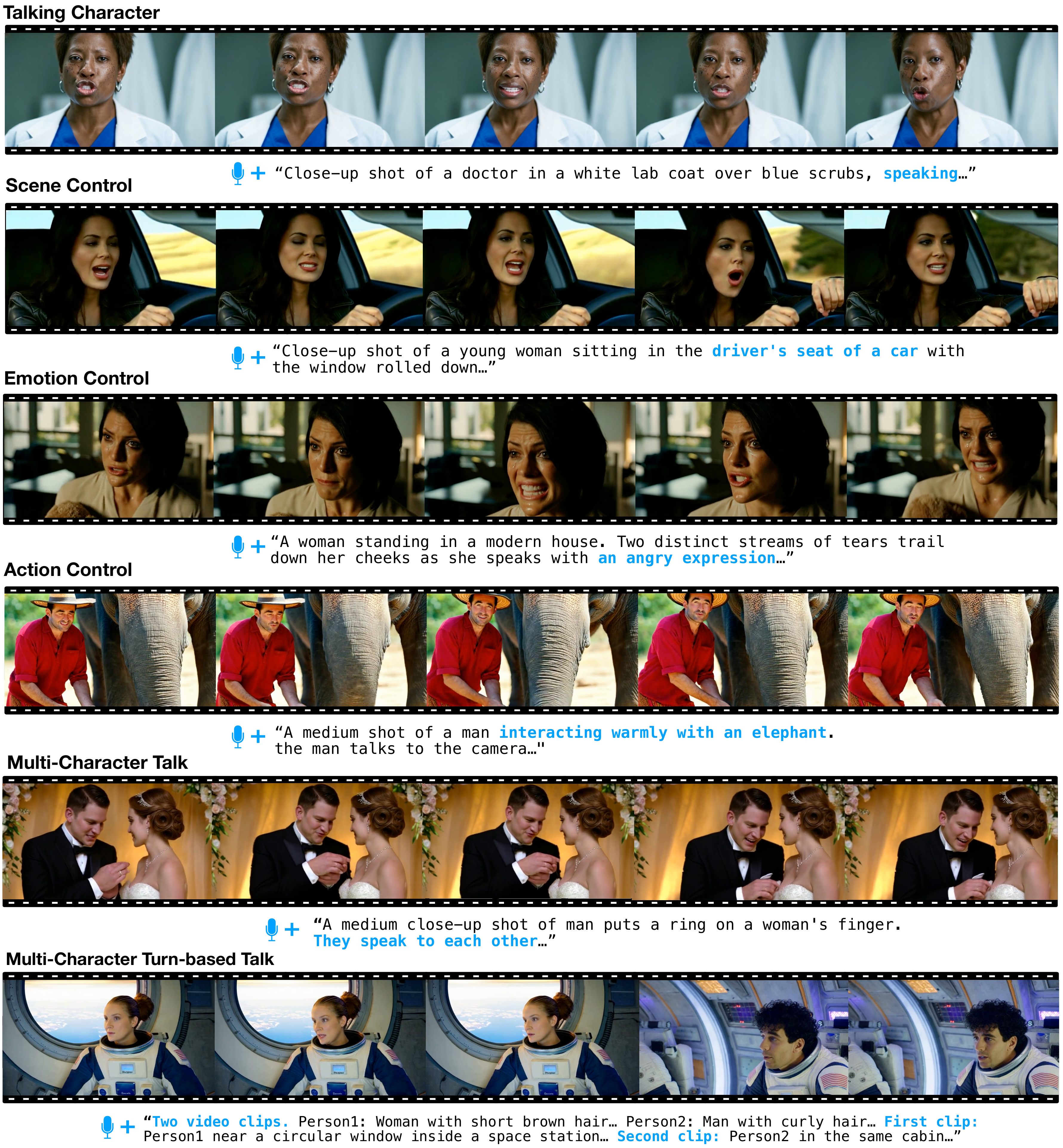}
    \caption{\MoCha is an end-to-end talking character video generation model that takes only \textbf{speech} and \textbf{text} as input, without requiring any auxiliary conditions. More videos are available on our website: \url{https://congwei1230.github.io/MoCha}}
    \label{fig:teaser}
    \vspace{-3pt}
\end{center}
}]


\begin{abstract}
Recent advancements in video generation have achieved impressive motion realism, yet they often overlook character-driven storytelling, a crucial task for automated film, animation generation. 
We introduce \textbf{Talking Characters}, a more realistic task to generate talking character animations directly from speech and text. Unlike talking head, Talking Characters aims at generating the full portrait of one or more characters beyond the facial region. 
In this paper, we propose MoCha, the first of its kind to generate talking characters. To ensure precise synchronization between video and speech, we propose a \textbf{speech-video window attention} mechanism that effectively aligns speech and video tokens.
To address the scarcity of large-scale speech-labeled video datasets, we introduce a joint training strategy that leverages both speech-labeled and text-labeled video data, significantly improving generalization across diverse character actions. We also design structured prompt templates with character tags, enabling, for the first time, \textbf{multi-character conversation with turn-based dialogue}—allowing AI-generated characters to engage in context-aware conversations with cinematic coherence.
Extensive qualitative and quantitative evaluations, including human preference studies and benchmark comparisons, demonstrate that MoCha sets a new standard for AI-generated cinematic storytelling, achieving superior realism, expressiveness, controllability and generalization.
\end{abstract}

\begin{figure*}[!ht]
    \centering
    \includegraphics[width=\linewidth]{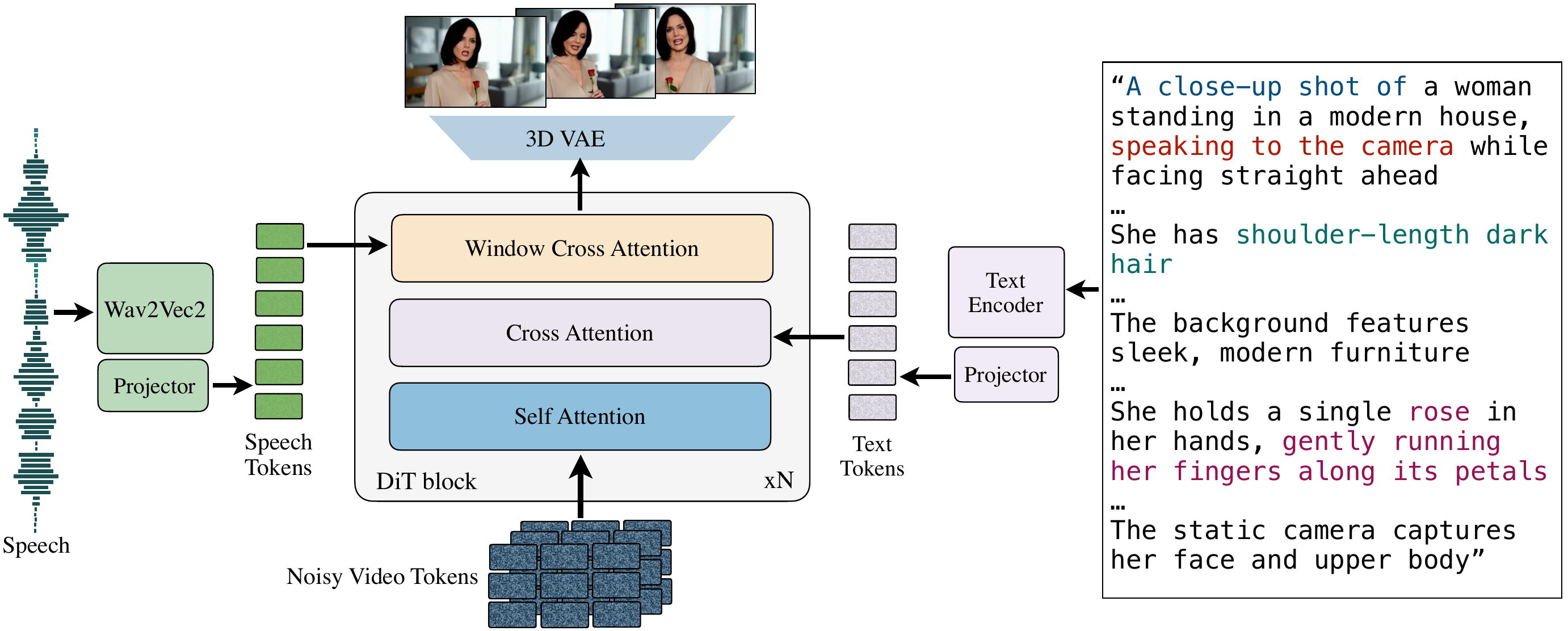}
    \caption{\textbf{MoCha Architecture.} \MoCha is a end-to-end Diffusion Transformer model that generates video frames from the joint conditioning of \textbf{speech} and \textbf{text}, without relying on any auxiliary signals. Both speech and text inputs are projected into token representations and aligned with video tokens through cross-attention.}
    \label{fig:framework}
\end{figure*}

\section{Introduction}
\label{sec:introduction}
Automating film production holds immense commercial potential, promising to democratize cinematic-level storytelling by enabling content creators to effortlessly generate films through natural language~\citep{zhou2024storydiffusion,mocogan,videodiffusion,videogpt}. Ideally, creators should be able to specify rich narratives involving multiple characters—either realistic humans or stylized cartoons—that engage in meaningful dialogues, expressive emotional portrayals, synchronized speech, and realistic full-body actions. Crucially, talking characters serve as powerful mediums for delivering impactful messages, clearly communicating ideas, and deeply engaging audiences. In films especially, dialogue serves as a key vehicle to effectively convey narratives, with vast downstream applications including digital assistants, virtual avatars, advertising, and educational content.

However, existing video foundation models are far from achieving this vision. Despite significant advancements in visually compelling content and dynamic environments, models such as SoRA, Pika, Luma, Hailuo, and Kling~\citep{chen2025goku,videocrafter1,yang2024cogvideox,polyak2024moviegencastmedia,stablevideodiffusion,kong2024hunyuanvideo,chefer2025videojam,brooks2024video} primarily generate characters with limited speech capabilities. Typically, these models exhibit simplified mouth movements and emotional expressions detached from meaningful dialogue, lacking control over actual spoken content. Consequently, their practical usability is severely restricted for speech-driven interactions essential to cinematic and interactive applications. On the other hand, recent speech-driven video generation methods, such as Loopy, Hallo3, and EMO~\cite{meng2024echomimicv2,emo,emo2,jiang2024loopy,xu2024hallo,tian2024emo}, predominantly focus on synthesizing talking-head videos confined to facial regions. These approaches neglect essential full-body movements and multi-character interactions critical for expressive storytelling, thus significantly limiting their applicability in realistic and interactive cinematic scenarios.

To bridge these gaps, we introduce the novel task: \textbf{Talking Characters}, defined as generating characters from natural language and speech input that naturally express synchronized speech, realistic emotions and full-body actions. We further propose \textbf{\MoCha}, the first-of-its-kind diffusion transformer (DiT) model trained end-to-end to achieve high-quality, movie-grade talking character generation.

\MoCha introduces several key technical innovations tailored specifically for this task:

\begin{itemize}

    \item \textbf{End-to-End Training Without Auxiliary Conditions}: Unlike prior works such as EMO~\citep{emo,emo2}, SONIC~\citep{sonic}, Echomimicv2~\citep{meng2024echomimicv2}, Loopy~\citep{jiang2024loopy}, and Hallo3~\citep{xu2024hallo}, which rely heavily on external control signals (e.g., reference images, skeletons, keypoints), \MoCha is trained directly on text and speech without any auxiliary conditioning. This simplifies the model architecture and improves motion diversity and generalization.
    
    \item \textbf{Speech-Video Window Attention}: We propose a novel attention mechanism that aligns speech and video inputs through localized temporal conditioning (see Sec.~\ref{sec:speech_video_window_attention}). This design significantly improves lip-sync accuracy and speech-video alignment.
    
    \item \textbf{Joint Speech-Text Training Strategy}: To address the scarcity of large-scale speech-labeled video datasets, we introduce a joint training framework that leverages both speech-labeled and text-labeled video data. This strategy enhances the model’s ability to generalize across diverse character actions and enables \textit{universal controllability} through natural language prompts, enabling nuanced control of character expressions, actions, interactions, and environments without auxiliary signals.
    
    \item \textbf{Multi-Character Conversation Generation}: For the first time, \MoCha enables coherent multi-character conversations in dynamic, turn-based dialogues, overcoming the single-character limitation of prior methods and supporting cinematic, story-driven video synthesis.
\end{itemize}

To evaluate \MoCha's performance, we curated \textbf{\MoChaBench}, a benchmark tailored for Talking Characters generation tasks. Both human evaluations and automatic metrics demonstrate that \MoCha set a new standard for talking character video generation and represents a significant step toward achieving controllable, narrative-driven video synthesis, with broad applications in film production, animation, virtual assistants, and beyond.

\begin{figure*}[!ht]
    \centering
    \includegraphics[width=0.7\linewidth]{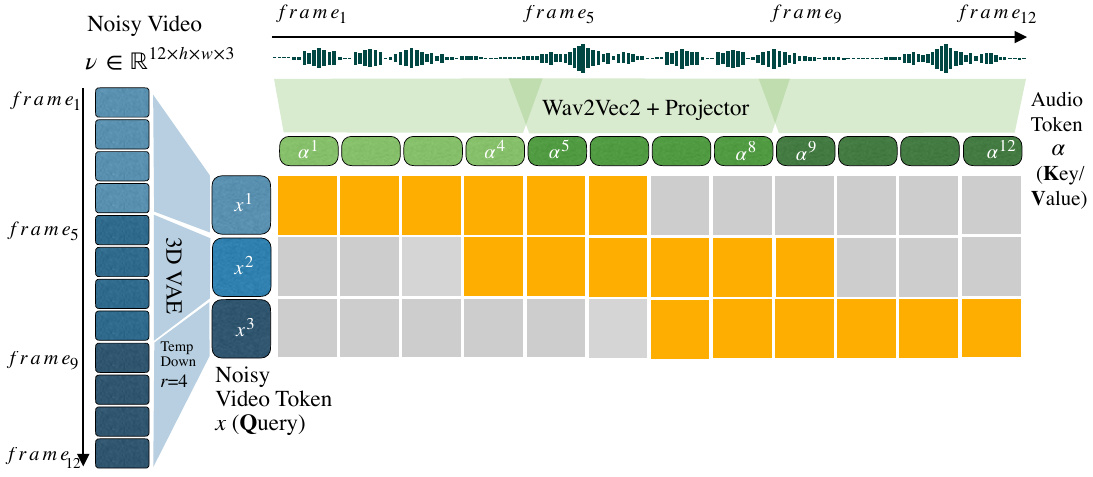}
    \caption{\textbf{\MoCha's Speech-Video Window Cross Attention} 
    \MoCha generates all video frames in parallel using a window cross-attention mechanism, where each video token attends to a local window of audio tokens to improve alignment and lip-sync quality.
    }
    \label{fig:window_attention}
\end{figure*}

\section{Task: Talking Characters}
\label{sec:talking_characters}
We introduce the novel task of Talking Characters, which aims to generating characters from natural language and speech input that mimicking realistic human-like behaviors. In contrast to talking-head (Close-up shot), Talking Characters aims at generating digital characters at any camera shot size (From close-up shot to wide-shot), covering one or more characters beyond the facial region. 

\noindent \textbf{Input:}
A Talking Character system takes as input:
\begin{enumerate}
    \item A \textit{text prompt} describing the character, environment, actions, facing direction(optional), position in the frame(optional) and camera framing(optional).
    \item A \textit{speech audio} for driving the character mouth, facial expression and body motion.
\end{enumerate}

\noindent \textbf{Output:}
The output is a video featuring one or more talking characters, which can be human, 3D cartoon, or animal.

\noindent \textbf{Evaluation:}
The generated characters are expected to perform well across the following five axes:
\begin{enumerate}
    \item Lip-Sync Quality: Speak the provided audio with accurate and temporally aligned lip synchronization.
    \item Facial Expression Naturalness: Express natural and coherent facial emotions that align with both the speech content and the text prompt.
    \item Action Naturalness: Perform natural and fluid body movements corresponding to the actions described in the text, with gestures synchronized to the speech.
    \item Text Alignment: Appear in a scene and context that are consistent with the descriptions provided in the prompt.
    \item Visual Quality: The entire video is visual consistency and temporal coherence without visual artifacts.
\end{enumerate}

\section{Model: \MoCha}
\label{sec:method}
In this section, we introduce the \MoCha model, the first model to
generate talking characters. We begin by outlining its architecture in \autoref{sec:joint_speech_text_dit}, followed by the speech-video window attention mechanism in \autoref{sec:speech_video_window_attention}. Next, we describe the method of generating multiple clips in \autoref{sec:multi_clip_generation}. Finally we provide explanation of the training strategy in \autoref{sec:training_strategy}.
\subsection{Speech\,+\,Text to Video Diffusion Transformers}
\label{sec:joint_speech_text_dit}

Figure~\ref{fig:framework} presents the overall framework of \MoCha. 
Unlike prior works that employ text-to-image (T2I) U-Net~\cite{emo, emo2, xu2024hallo, cui2024hallo2} for talking head generation, \MoCha is built on a diffusion transformer (DiT)~\cite{Peebles2022DiT}. By incorporating text and speech conditions sequentially via cross-attention, it effectively captures both semantics and temporal dynamics.

\paragraph{Model Architecture.}
Given an RGB video \(\nu \in \mathbb{R}^{T \times H \times W \times 3}\) with \(T\) frames, we encode it into a latent representation 
\(
  x_0 \in \mathbb{R}^{\tau \times h \times w \times c}
\) 
using a 3D VAE, which down-samples the video spatially and temporally. We define the temporal down-sampling ratio as 
\(
  r = \frac{T}{\tau}.
\)
Next, \(x_0\) is flattened into a sequence of tokens of size 
\(
  (\tau \times h \times w) \times c
\)
and passed to the DiT model \(f_\theta(\cdot)\). 
Within each DiT block, the model first applies self-attention to the tokens, followed by sequential cross-attention with the text condition tokens \(c\) and audio condition tokens \(\alpha\). The audio condition \(\alpha \in \mathbb{R}^{T \times c}\) is derived from raw waveforms using Wav2Vec2~\cite{baevski2020wav2vec} and processed through an single layer MLP to align its feature dimension with the latent video tokens.

\paragraph{Training Objective.}
We adopt Flow Matching~\cite{lipman2022flow}, which enables efficient simulation of continuous-time dynamics, to train our model. Given a latent video representation \(x_1 \in \mathbb{R}^{\tau \times h \times w \times c}\) (encoded from the input video), random noise \(\epsilon \sim \mathcal{N}(0,I)\), and a continuous time step \(t \in [0,1]\), we construct an intermediate latent \(x_t\) by interpolating between \(\epsilon\) and \(x_1\):
\begin{equation}
    x_t = (1 - t) \, \epsilon + t \, x_1.
\end{equation}
The model is trained to predict the velocity, defined as the difference between the data and noise:
\begin{equation}
    v_t = \frac{d x_t}{d t} = x_1 - \epsilon.
\end{equation}
The training loss is then:
\begin{equation}
    L = \mathbb{E}_{\epsilon \sim \mathcal{N}(0,I),\, x_1,\, c,\, \alpha,\, t \in [0,1]}
    \Bigl\| f_\theta\bigl(x_t, c, \alpha, t\bigr) - (x_1 - \epsilon) \Bigr\|_2^2,
\end{equation}
where \(x_1\) is the encoded latent video, \(c\) and \(\alpha\) are text and audio conditions, and \(f_\theta(\cdot)\) is the DiT model.

\subsection{Speech-Video Window Attention}
\label{sec:speech_video_window_attention}
Most talking head generation methods employ 2D diffusion models (e.g., U-Net) that auto-regressively generate $T$ video frames conditioned on audio tokens $\alpha \in \mathbb{R}^{T \times c}$.
When generating frame $\nu_i$, the model is provided only with the corresponding audio token $\alpha_i$.
This design inherently preserves speech-video synchronization, ensuring accurate alignment between lip movements and the corresponding speech.
However, when using DiT-style architectures, two key differences emerge that disrupt this alignment:
\begin{enumerate}
    \item Temporal Compression: Videos are compressed using a 3D VAE with a downsampling ratio $r$ (typically $r=4$ or $8$ in modern T2V models~\cite{polyak2024moviegencastmedia,kong2024hunyuanvideo}), resulting in latent representations of length $\tau = T/r$. While audio remains at the original resolution ($T$), video tokens operate at the compressed scale ($\tau$), degrading lip synchronization.
    \item Parallel Generation: Unlike autoregressive models, DiT generates all $\tau$ latent frames in parallel. However, naïve cross-attention allows each video token to attend to all audio tokens. As a result, latent frames may incorrectly associate with phonemes from unrelated timesteps.
\end{enumerate}

To address this, we propose a \textit{Speech-Video Window Attention} mechanism that enforces localized conditioning. This design is motivated by the observation that lip movements depend on short-term audio cues (1–2 phonemes), whereas body motions align with longer-term text descriptions. To capture this distinction, we constrain each video token to attend only to a temporally bounded audio window.
As illustrated in Figure~\ref{fig:window_attention}, for each latent video frame $x^{(i)} \in \mathbb{R}^{h \times w \times c}$ ($i \in \{1,\dots,\tau\}$), attention is computed over audio tokens $\alpha^{j}$, where $j$ spans:
\begin{equation}
    j \in \left[ \max(1, (i-1)r - 1), \min(T, ir + 1) \right].
\end{equation}
This window encompasses $r+2$ audio tokens, covering the $r$ frames corresponding to the latent $x^{(i)}$, plus one token on either side to ensure contextual continuity, thereby enhancing local smoothness between adjacent latents.

\begin{figure*}[!ht]
    \centering
    \includegraphics[width=\linewidth]{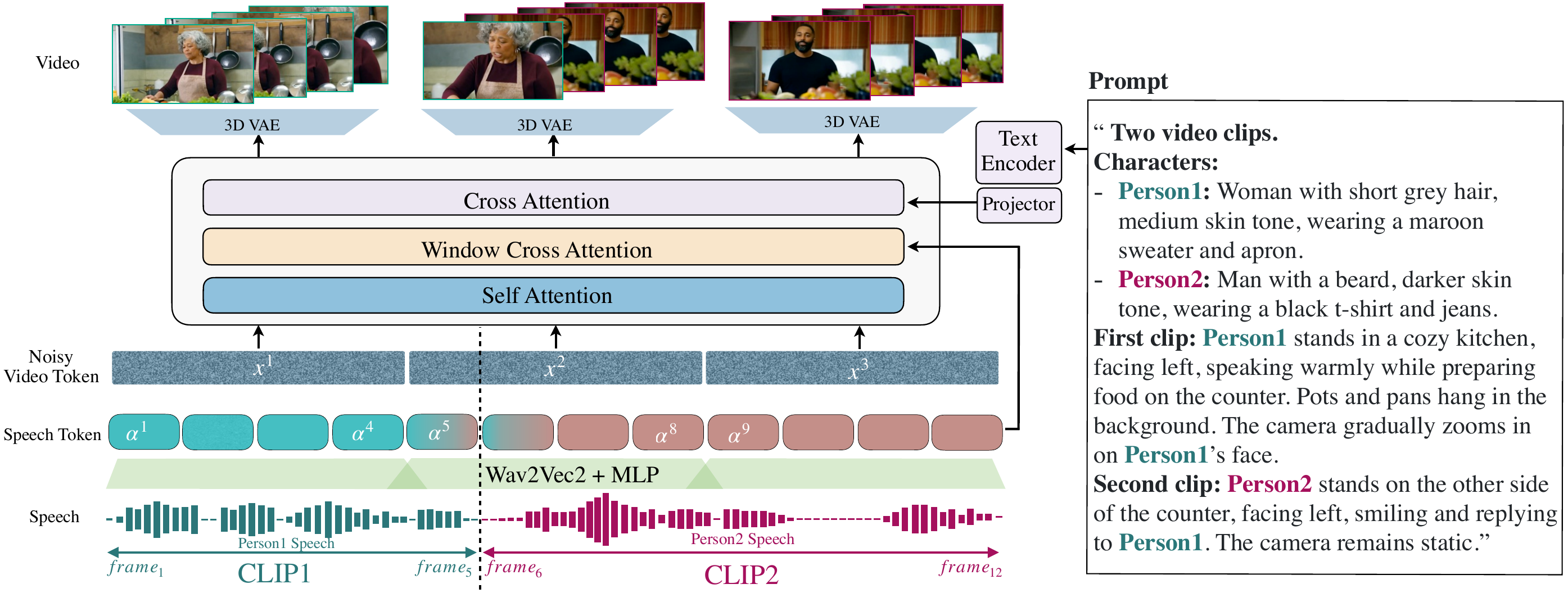}
\caption{\textbf{Multi-character Conversation and Character Tagging.} \MoCha supports generates multi-character conversion with scene cut. We design a specialized prompt template: it first specifies the number of clips, then introduces the characters along with their descriptions and associated tags. Each clip is subsequently described using only the \textbf{character tags}, simplifying the prompt while preserving clarity. \MoCha leverages self-attention across video tokens to ensures character and environment consistency. The audio conditioning signal implicitly guides the model on when to transition between clips.
}
    \label{fig:multi_clip_generation}
\end{figure*}

\subsection{Multi-character Conversation}
\label{sec:multi_clip_generation}
\MoCha generates multi-clip videos simultaneously in the same manner as single-clip generation, with no additional architectural modifications. As illustrated in Figure~\ref{fig:multi_clip_generation}, instead of auto-regressive generation in video extension methods which requiring conditioning on previous generated results, \MoCha leverages self-attention across video tokens to ensure consistency of characters appearing in multiple clips, as well as coherence in the surrounding environment. We assume that only one character speaks at a single time, so the changes in the speaker in the audio conditioning implicitly guide the \MoCha on when to transition between clips without any additional guidance signal condition such as clip tokens~\cite{wu2025mint}.

Binding attributes and actions to the correct characters using only text is particularly challenging in multi-clip settings—especially when multiple characters interact or when the same character appears across clips. Naive captioning models typically rely on visual descriptions to refer to characters. As a result, they must repeat detailed appearance descriptions each time a character is mentioned, leading to long, redundant, and confusing prompts. For example:
\begin{quote}
\small
\texttt{The girl aged 10-15 in a yellow dress waves to another girl dressed in a green shirt with braided hair holding a book... The girl in the green shirt with braided hair responds with a smile... Nearby, another girl aged 10-15 in a blue hoodie points toward the whiteboard while looking at the girl dressed in a green shirt with braided hair...}
\end{quote} 
This verbosity not only increases the risk of exceeding token limits (e.g., 256 tokens) but also confuses the model during generation—especially in multi-clip scenarios.

As shown in Figure~\ref{fig:multi_clip_generation}, we address this by introducing a structured prompt template with fixed keywords and a character tagging mechanism that promotes clarity, compactness, and consistency:

\begin{itemize}
    \item ``Two video clips" Specifies the number of clips.
    \item ``Characters" Introduces a list of characters, each described by visual attributes and assigned a unique tag \cite{liang2025movie} (e.g., \texttt{Person1}, \texttt{Person2}).
    \item ``First clip", ``Second clip", etc. Each video segment is described using only the defined character tags.
\end{itemize}
This design significantly reduces redundancy and helps the model reliably associate visual attributes with character actions, even across multiple clips.

\begin{figure*}[!ht]
    \centering
    \includegraphics[width=\textwidth]{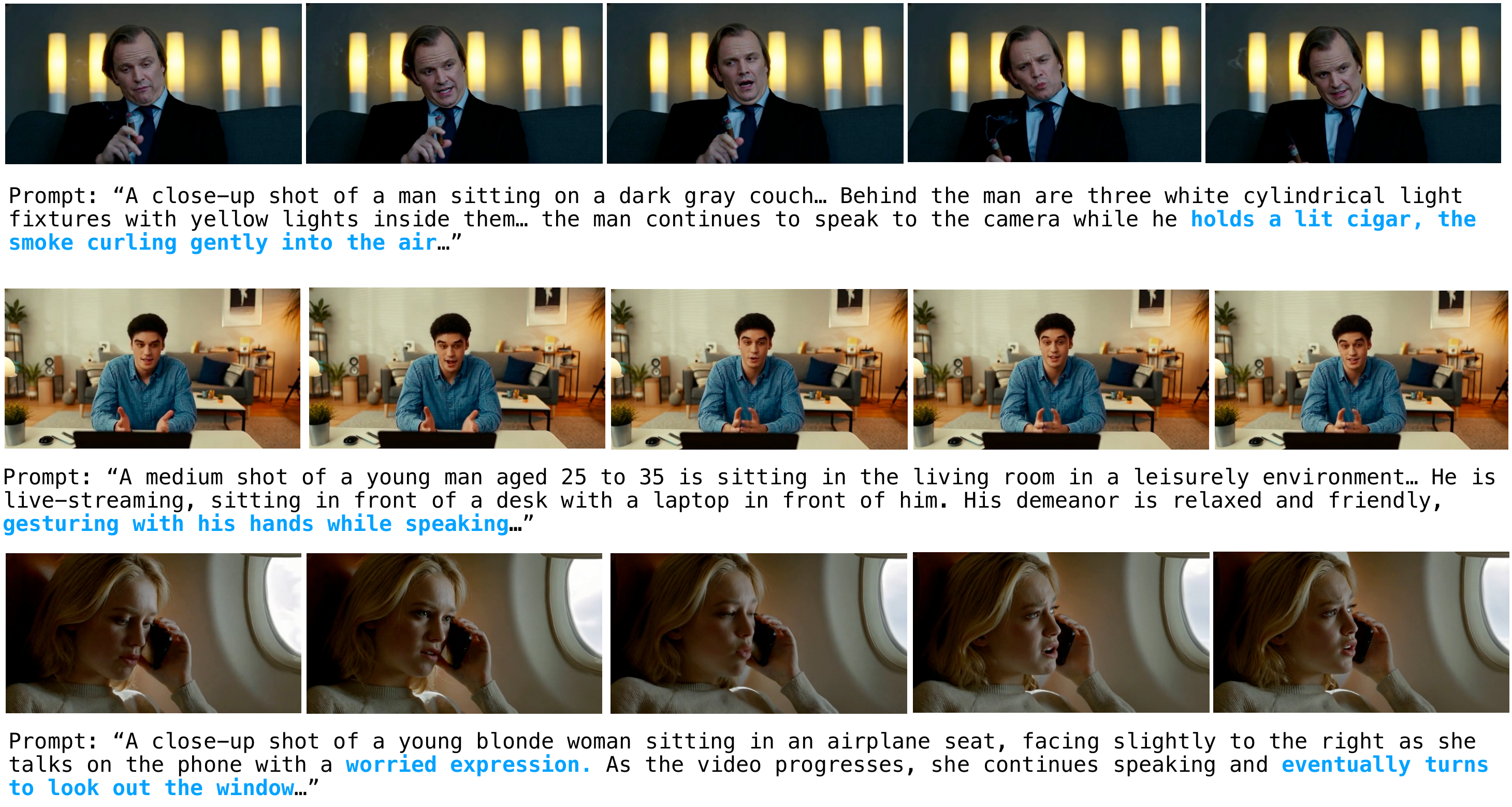}
        \caption{\textbf{Qualitative results of \MoCha on \MoChaBench}.
\MoCha not only generates lip movements that are well-synchronized with the input speech, but also produces natural facial expressions that reflect the prompt along with realistic hand gestures and action movements}
    \label{fig:MoCha_bench_demo}
\end{figure*}

\begin{figure}[h!]
    \includegraphics[width=\columnwidth]{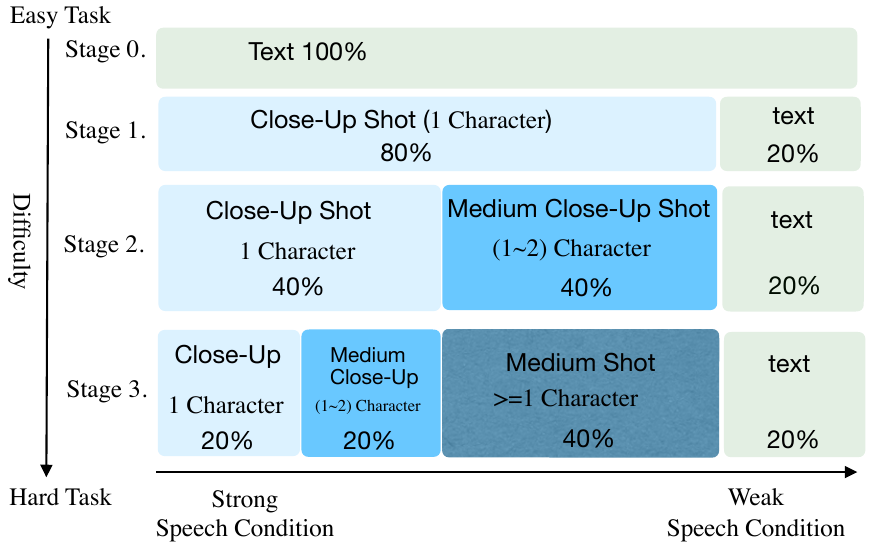}
    \caption{\textbf{Multi-Stage Training Strategy for \MoCha.} Text-Speech Joint training starts with close-up shots where speech conditioning has the strongest influence. At each stage, previous data is reduced by 50\%, and harder tasks with weaker speech conditioning are introduced. Stage 0 uses text-only video data to establish a foundation for the future stages.}
    \label{fig:training_strategy}
\end{figure}

\subsection{\MoCha Training Strategy}
\label{sec:training_strategy}
\noindent \textbf{Joint Training of (Speech+Text)-to-Video (ST2V) and Text-to-Video (T2V)}  
Speech-annotated video datasets are significantly smaller in scale and less diverse compared to standard text-to-video (T2V) datasets, making it challenging to train high-quality (Speech+Text)-to-Video (ST2V) models directly. Relying solely on speech-annotated data limits the model's ability to generalize across varied visual and semantic contexts. To address this, we propose a joint training strategy that integrates both speech-annotated and text-only video dataset:  
\begin{itemize}
    \item \textbf{80\% ST2V Data:} The model is primarily trained on speech-conditioned video data, leveraging both speech and text modalities.
    \item \textbf{20\% T2V Data:} To enhance diversity, we incorporate text-only video data, where speech conditioning is absent. In these cases, Wav2Vec2 embeddings are replaced with zero vectors, allowing the model to generalize across text-only prompts after training.
\end{itemize}


\noindent \textbf{Multi-Stage Human Video Learning}  
Speech conditioning in human video generation exhibits a diminishing influence as we progress from low-level to high-level motion: it strongly governs lip movements and facial expressions, but its effect weakens for co-speech gestures and full-body actions. Meanwhile, generating these higher-level motions is inherently more difficult. As a result, training on all types of speech-conditioned data simultaneously can lead to inefficiencies.

As illustrated in Figure~\ref{fig:training_strategy}, we address this challenge with a multi-stage training framework that categorizes data based on shot types, ranging from close-up to medium shots:
\begin{itemize}
    \item We begin training with close-up shots, which contain the strongest speech-video correlation.  
    \item At each subsequent stage, we reduce the data from the previous stage by 50\% while introducing harder tasks with weaker speech conditioning.  
    \item We maintain the 80\% ST2V and 20\% T2V data ratio across all stages to ensure balanced training.  
\end{itemize}  
In Stage 0, we pretrain \MoCha exclusively on text-conditioned video data (T2V) to establish a strong foundational prior for video generation before incorporating speech-conditioning signals.


\section{Experiment}
\label{sec:experiment}

In this section, we first describe our training data processing pipeline in \autoref{sec:training_data_processing_pipeline}, followed by the details of our model in \autoref{sec:experiment_model_detail}. We then introduce \MoChaBench for Talking Characters task and benchmark \MoCha against baseline methods in \autoref{sec:experiment_evaluation}, and finally, we present an ablation study to analyze the impact of key design choices in \autoref{sec:experiment_ablation_studies}.

\begin{figure*}[!ht]
    \centering
    \includegraphics[width=0.92\textwidth]{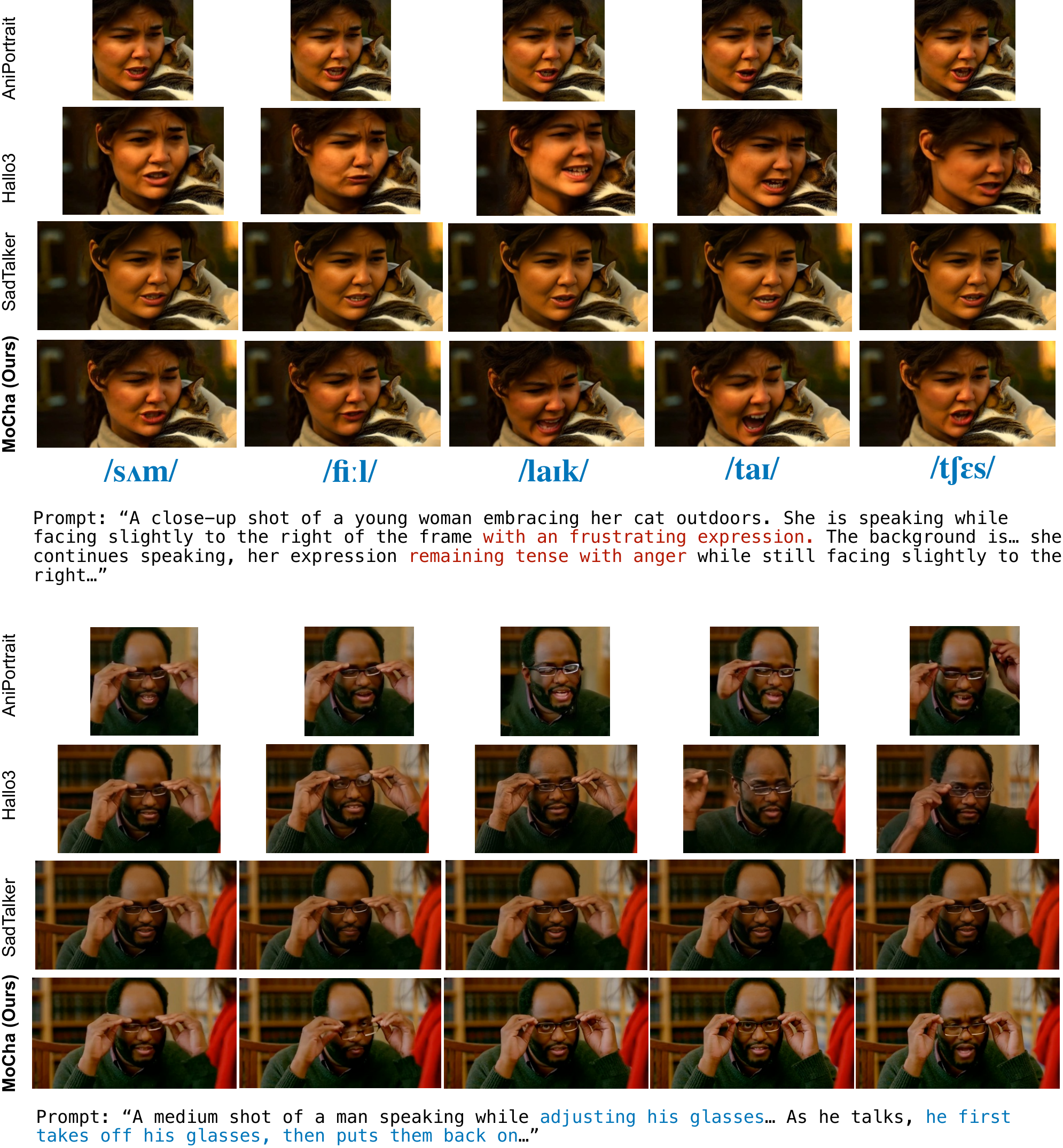}
    \caption{\textbf{Qualitative comparison between \MoCha and baselines on \MoChaBench.}
    \MoCha not only produces lip movements that align closely with the input speech—enhancing the clarity and naturalness of articulation—but also generates expressive facial animations and realistic, complex actions that faithfully follow the textual prompt. In contrast, SadTalker and AniPortrait exhibit minimal head motion and limited lip synchronization. Hallo3 mostly follows the lip-syncing but suffers from inaccurate articulation, erratic head movements, and noticeable visual artifacts.
        Since the baselines operate in an image-to-video (I2V) setting, we provide them with the first frame generated by \MoCha as input for comparison. The first frame is cropped and resized as needed to meet the requirements of each baseline.
    }
    \label{fig:comparison_with_baseline}
\end{figure*}

\begin{figure*}[h!]
    \includegraphics[width=\linewidth]{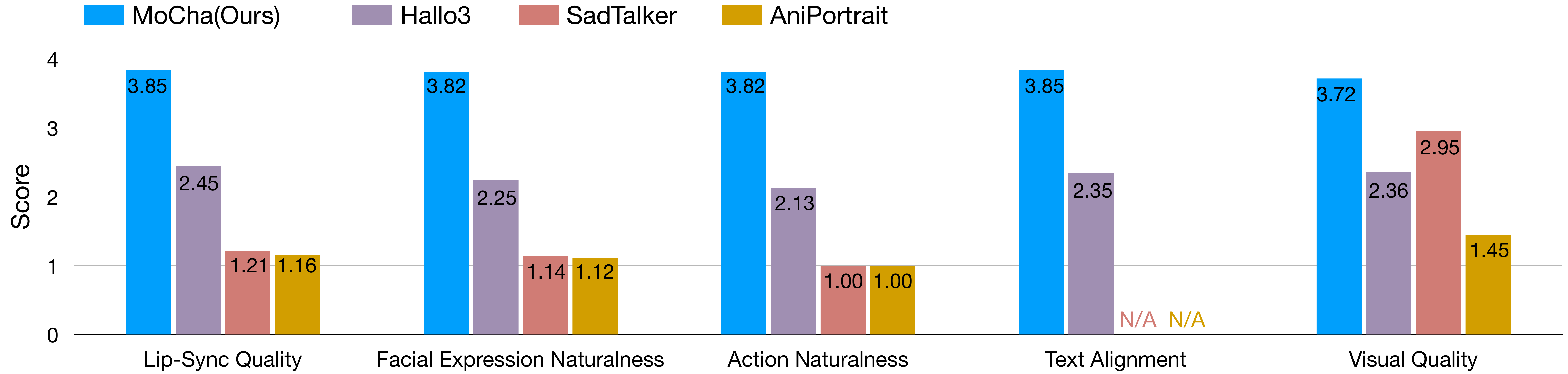}
    \caption{\textbf{Human evaluation scores on \MoChaBench.}
    Scores range from 1 to 4 across five evaluation axes, where a score of 4 reflects performance that is nearly indistinguishable from real video or cinematic production. \MoCha significantly outperforms all baselines across all axes. SadTalker and AniPortrait consistently received a score of 1 for action naturalness, as these methods only perform head movements. Text alignment is marked as not applicable (N/A) for these baselines since they do not accept text input.}
    \label{fig:MoCha_bench_results}
\end{figure*}

\subsection{Training Data Processing Pipeline}
\label{sec:training_data_processing_pipeline}

To construct high-quality training data, we employ a multi-stage filtering and annotation pipeline.

\begin{itemize}
    \item \textbf{Speech Scene Filtering:} We first segment videos into scenes using PySceneDetect~\cite{Castellano_PySceneDetect}. Each scene undergoes speech detection, where non-speech segments and those heavily influenced by background noise or music are discarded.  For each valid segment, we perform music and noise removal and Wav2Vec2~\cite{schneider2019wav2vec,baevski2020wav2vec} is used to extract speech embeddings for valid segments.
    
    \item \textbf{Prominent Character Filtering:} To ensure the dataset focuses on scenes with clear, prominent human characters, we employ an LLM-based filtering mechanism. The model analyzes each scene and removes those lacking a central characters.

    \item \textbf{Motion and Lip-Sync Filtering:} We further refine the dataset by applying motion and lip-sync filters, ensuring that the extracted speech corresponds to meaningful human expressions and actions.

    \item \textbf{Scene Captioning:} Each processed scene is captioned using an LLM~\cite{meta2024introducing}, describing character appearances, positions, speech activity, emotions, and body language in highly detailed structured format.
\end{itemize}
Following this pipeline, we curate a high-quality dataset consisting of 300 hours (\(\mathcal{O}(500)K\) samples) of speech-conditioned video data.

\subsection{Implementation Details}
\label{sec:experiment_model_detail} 
Our model follows a design similar to MovieGen~\cite{polyak2024moviegencastmedia} and HunyuanVideo~\cite{kong2024hunyuanvideo}, utilizing a DiT architecture. \MoCha is built on a pretrained 30-B DiT model. All models are trained with a spatial resolution of approximately $720 \times 720$, accommodating multiple aspect ratios. The model is optimized to generate 128-frame videos at 24 frames per second, resulting in outputs with a duration of 5.3 seconds. Our dataset~\cite{cui2024hallo3} for text conditioning comprises approximately $\mathcal{O}(100)M$ samples, while the speech conditioning video dataset consists of around $\mathcal{O}(500)K$ samples. Training is conducted on 64 nodes to support the scale of our model.

\begin{table*}[t]
    \centering
    \footnotesize
    \begin{tabular}{lccccc}
        \toprule
        \textbf{Method} & \textbf{Lip-Sync Quality} & \textbf{Facial Expression Naturalness} & \textbf{Action Naturalness} & \textbf{Text Alignment} & \textbf{Visual Quality} \\
        \midrule
        Hallo3~\cite{xu2024hallo} & \underline{2.45} & \underline{2.25} & \underline{2.13} & \underline{2.35} & \underline{2.36} \\
        SadTalker~\cite{zhang2023sadtalker} & 1.21 & 1.14 & 1.00 & N/A & 2.95 \\
        AniPortrait~\cite{wei2024aniportrait} & 1.16 & 1.12 & 1.00 & N/A & 1.45 \\
        \midrule
        \textbf{MoCha (Ours)} & \textbf{3.85} \textcolor{darkgreen}{(+1.40)} & \textbf{3.82} \textcolor{darkgreen}{(+1.57)} & \textbf{3.82} \textcolor{darkgreen}{(+1.69)} & \textbf{3.85} \textcolor{darkgreen}{(+1.50)} & \textbf{3.72} \textcolor{darkgreen}{(+1.36)} \\
        \bottomrule
    \end{tabular}
    \caption{\textbf{Human evaluation scores on MoCha-Bench.} Scores range from 1 to 4 across five evaluation axes, where a score of 4 reflects performance that is nearly indistinguishable from real video or cinematic production. Participants rated each method on five aspects: lip-sync quality, facial expression naturalness, action naturalness, text-prompt alignment, and visual quality. MoCha significantly outperforms prior methods across all categories. Green numbers indicate absolute improvements ($\Delta$) over the second-best method (underlined). SadTalker and AniPortrait consistently received a score of 1 for action naturalness, as these methods only perform head movements. }
    \label{tab:human_eval}
\end{table*}

\begin{table}[t]
    \centering
    \small
    \begin{tabular}{lcc}
        \toprule
        \textbf{Method} & \textbf{Sync-C} $\uparrow$ & \textbf{Sync-D} $\downarrow$ \\
        \midrule
        SadTalker~\cite{zhang2023sadtalker} & 4.727 & 9.239 \\
        AniPortrait~\cite{wei2024aniportrait} & 1.740 & 11.383 \\
        Hallo3~\cite{xu2024hallo} & \underline{4.866} & \underline{8.963} \\
        \textbf{Ours} & \textbf{6.037} \textcolor{darkgreen}{(+1.17)} & \textbf{8.103} \textcolor{darkgreen}{(-0.86)} \\
        \bottomrule
    \end{tabular}
    \caption{\textbf{Comparison with State-of-the-Art Methods on \MoChaBench.} We report synchronization metrics: Sync-C (higher is better) and Sync-D (lower is better). \MoCha outperforms all baselines, indicating superior lip-sync quality.}
    \label{tab:benchmark_comparison}
\end{table}

\begin{table}[t]
    \centering
    \resizebox{1.0\linewidth}{!}{
    \begin{tabular}{lcc}
        \toprule
        \textbf{Ablation} & \textbf{Sync-C} $\uparrow$ & \textbf{Sync-D} $\downarrow$ \\
        \midrule
        \textbf{Ours} & \textbf{6.037} & \textbf{8.103} \\
        \midrule
        \textbf{w/o Joint ST2V + T2V Training} & 5.659 & 8.435 \\
        \textbf{w/o Speech-Video Window Attention} & 5.103 & 8.851 \\
        \bottomrule
    \end{tabular}
    }
        \caption{\textbf{Ablation Study of \MoCha on \MoChaBench} We analyze the impact of different components by disabling them and measuring the effect on key metrics. Removing speech-video window attention degrades synchronization, joint ST2V and T2V training improves generalization.}
    \label{tab:ablation}
\end{table}

\subsection{Evaluation}
\label{sec:experiment_evaluation}

\noindent \textbf{Baselines.}
We compare our approach against several representative audio-driven talking face generation methods with publicly available source code or implementations, including SadTalker~\cite{zhang2023sadtalker}, AniPortrait~\cite{wei2024aniportrait}, and Hallo3~\cite{xu2024hallo}. These baselines span both end-to-end architectures and those utilizing intermediate facial representations, enabling a comprehensive evaluation of our method across diverse generation paradigms.

\noindent \textbf{Benchmark.}
We introduce \textbf{MoCha-bench}, a benchmark tailored for the Talking Character generation task. It comprises 150 diverse examples, each consisting of a text prompt and corresponding audio clip. The dataset includes both close-up and medium-shot compositions: close-ups emphasize facial expressions and lip synchronization, while medium shots highlight hand gestures and broader body movements. The scenes span a wide range of human activities and interactions with objects (e.g., \texttt{a chef chopping vegetables}, \texttt{a musician playing an instrument}) and the character speaking with various emotions and facing directions.
All text prompts were manually curated and further enriched using the publicly released LLaMA-3~\cite{dubey2024llama} model to enhance expressiveness and variety.
As \MoCha directly generates videos from speech and text inputs, while all baseline models operate in an image-to-video (I2V) setting, we ensure a fair comparison by providing each I2V method with the first frame of the \MoCha's generation as the input.

\noindent \textbf{Qualitative Experiments}
We presents qualitative results of \MoCha-30B in \cref{fig:teaser} and \cref{fig:MoCha_bench_demo}  showcasing its ability to generate diverse and realistic human motion while synchronizing speech with complex actions.

\cref{fig:comparison_with_baseline} presents a direct comparison between \MoCha and baseline methods on \MoChaBench. All baselines require a reference image as an auxiliary input. To ensure fairness, we first generate a video using \MoCha and then use its first frame as the reference image for all baseline models. For models that do not support arbitrary aspect ratios, we crop the first frame to focus on the head region before feeding it into their networks.
We provide two groups of qualitative comparisons: one featuring close-up shots and the other medium shots. The close-up group emphasizes lip-sync quality, head movement, and facial expressions, while the medium shot group focuses on hand movements during speech.
\MoCha not only produces lip movements that closely align with the input speech—enhancing both articulation and naturalness—but also generates expressive facial animations and realistic, coordinated actions that accurately follow the textual prompt. In contrast, SadTalker and AniPortrait exhibit minimal head motion and limited lip synchronization. While Hallo3 achieves mostly consistent lip-syncing, it suffers from inaccurate articulation and erratic head movements. In the medium shot comparisons, Hallo3 also introduces noticeable visual artifacts, particularly during complex actions.

\noindent \textbf{Quantitative Experiments}
We evaluate video quality using the automatic metrics to measure the lip-sync quality.
Table~\ref{tab:benchmark_comparison} presents a comparison on the \MoChaBench.
Our model achieves the best scores across lip-sync metrics.

\noindent \textbf{Human Evaluations.}  
We conduct a comprehensive human evaluation to compare \MoCha against baseline methods on the \MoChaBench dataset. The evaluation is based on five axes tailored for the Talking Characters task(See \ref{sec:talking_characters}):

\begin{itemize}
    \item \textbf{Lip-Sync Quality:} Measures how accurately the character’s lip movements align with the spoken audio.\\
    \textit{Scale:} 
    1 – Not aligned at all, 
    2 – Weak alignment, 
    3 – Mostly aligned, 
    4 – Perfectly aligned.

    \item \textbf{Facial Expression Naturalness:} Evaluates whether the facial expressions and lip-sync appear natural and contextually coherent, without seeming robotic or exaggerated.\\
    \textit{Scale:}
    1 – Completely unnatural, 
    2 – Noticeably synthetic or stiff, 
    3 – Mostly natural and believable, 
    4 – Indistinguishable from real or cinematic performance.

    \item \textbf{Action Naturalness:} Assesses how naturally the character's body movements and gestures align with the audio.\\
    \textit{Scale:}
    1 – Completely unnatural, 
    2 – Noticeably unnatural, 
    3 – Mostly natural, 
    4 – Indistinguishable from real movie or TV characters.

    \item \textbf{Text Alignment:} Measures how well the generated actions and expressions follow the behaviors described.\\
    \textit{Scale:}
    1 – No alignment, 
    2 – Partial alignment, 
    3 – Mostly aligned, 
    4 – Perfect alignment with the prompt.

    \item \textbf{Visual Quality:} Evaluates visual quality by checking for issues such as artifacts, discontinuities, or glitches.\\
    \textit{Scale:}
    1 – Severe artifacts, 
    2 – Noticeable artifacts, 
    3 – Mostly artifact-free, 
    4 – Flawless visuals.
\end{itemize}

Each model output received 5 independent ratings per example, resulting in over 750 responses per model. \MoCha significantly outperforms all baselines across all five axes, with average scores approaching 4—indicating performance that is nearly indistinguishable from real video or cinematic production.

\subsection{Ablation Studies}
\label{sec:experiment_ablation_studies}
We conduct ablation studies to analyze the contribution of key components in \MoCha. Table~\ref{tab:ablation} presents the impact of each design choice.

\begin{itemize}
    \item \textbf{Speech-Video Window Attention Ablation}: We disable our speech-video window attention mechanism to analyze its effect on speech-video alignment. This results in a noticeable drop in Sync-C (6.037 → 5.103) and increased Sync-D (8.103 → 8.851), confirming that our method significantly enhances lip synchronization.
    
    \item \textbf{Joint ST2V and T2V Training Ablation}: We train \MoCha exclusively on ST2V data (removing text-only video training). This leads to This results in a noticeable drop in Sync-C (6.037 → 5.659) and increased Sync-D (8.103 → 8.435), indicating degraded generalization due to reduced dataset diversity.    
\end{itemize}

These findings confirm that both our speech-video window attention and joint training strategy are essential for achieving high-quality motion, realistic speech alignment, and overall superior generation performance.

\section{Related Work}
\label{sec:related_work}
\subsection{Talking Head Generation}
Given an audio sequence and a reference face, pioneer talking-head generation works typically utilize biometric signals such as facial keypoints ~\cite{prajwal2020lip, shen2023difftalk, liu2023moda, zhong2023identity}, or 3D priors ~\cite{doukas2021headgan, ma2023dreamtalk, ji2021audio, sun2023vividtalk, gan2023efficient, zhang2023sadtalker, ye2024real3d} as intermediate motion representation to animate the reference face while ensuring lip synchronization. For example, SadTalker ~\cite{zhang2023sadtalker} first extracts 3DMM coefficients from audio and then renders the face in a 3D-aware manner. 
AniPortrait ~\cite{wei2024aniportrait} predicts 2D facial landmarks from audio and then utilizes a diffusion models to generate a portrait video from the 2D landmarks maps. VLOGGER \cite{corona2024vlogger} predicts both 3D expression coefficient and 3D body pose from speech and enables the simultaneous generation of talking-face animations and upper-body gestures. Although effective, videos generated by these methods often lack expressiveness and naturalness due to the limited representation of 2D/3D priors. 

Recently works, such as EMO \cite{tian2024emo} and Hallo \cite{xu2024hallo}, generate audio-driven portrait videos end-to-end using diffusion models, which eliminate intermediate facial representations and learn natural motion from data \cite{tian2024emo, xu2024hallo, cui2024hallo2, jiang2024loopy, wang2024v}. 
Hallo3 \cite{cui2024hallo3} builds upon pretrained transformer-based video diffusion models to animate faces with dynamic head poses and background elements. Although these methods can generate natural expressions, they rely on complex auxiliary signals—such as reference images or keypoints—which not only limit the naturalness and flexibility of facial expressions and body movements but also limit the generalization ablity of those methods.

\subsection{Diffusion-Based Video Generation}
Diffusion models have emerged as a powerful approach for video synthesis, demonstrating state-of-the-art results in text-to-video generation. Methods like Make-A-Video~\cite{singer2022make}, MagicVideo~\cite{zhou2022magicvideo}, and AnimateDiff~\cite{guo2023animatediff} leverage pre-trained text-to-image (T2I) models and extend them to the temporal domain to synthesize coherent motion sequences. Recent advances in DiT-based architectures, such as CogVideoX~\cite{yang2024cogvideox} and MovieGen~\cite{polyak2024moviegencastmedia}, have further improved video fidelity and controllability by integrating spatial and temporal constraints.

Despite these advancements, existing diffusion-based methods primarily focus on scene dynamics and global motion synthesis, lacking explicit modeling of speech-driven facial and body gestures of characters in the video. Our proposed \MoCha framework extends diffusion models to jointly condition on speech and text, enabling the generation of lifelike character animations with natural conversation.

\section{Conclusion}
\label{sec:conclusion}

In summary, our work pioneers the task of Talking Characters Generation, pushing beyond traditional talking head synthesis to enable full-body, multi-character animations directly driven by speech and text. We present \MoCha, the first framework to address this challenging task, introducing key innovations such as the speech-video window attention mechanism for precise audio-visual alignment and a joint training strategy that leverages both speech- and text-labeled data for enhanced generalization. Additionally, our structured prompt design unlocks multi-character, turn-based dialogues with contextual awareness, marking a significant step toward scalable, cinematic AI storytelling. Through comprehensive experiments and human evaluations, we demonstrate that MoCha delivers state-of-the-art performance in terms of realism, expressiveness, and controllability, setting a solid foundation for future research in generative character animation.

\section{Acknowledgment}
\label{sec:Acknowledgment}
We thank Xinyi Ji, Tianquan Di, Anqi Xu, Matthew Yu, Emily Luo for providing speech samples used in the \MoCha{} demo.

{
    \small
    \bibliographystyle{ieeenat_fullname}
    \bibliography{main}
}

\end{document}